\documentclass[sort&compress,preprint,12pt]{elsarticle}
\usepackage{amssymb}
\usepackage{amsmath}
\usepackage{algorithm}
\usepackage{algorithmicx}
\usepackage{algpseudocode}
\usepackage{booktabs}
\usepackage{caption}
\usepackage{textcomp}
\usepackage{url}

\journal{xxxx}

\begin{document}

\begin{frontmatter}
    \title{AIAuditTrack: A Framework for AI Security System} 
    \author[hust]{{Zixun Luo}}
    \author[ln]{Yuhang Fan}
    \author[CAIR]{Yufei Li}
    \author[CAIR]{Youzhi Zhang}
    \author[thu]{Hengyu Lin}
    \author[JX] {Ziqi Wang}

\affiliation[hust]{organization={Huazhong University of Science and Technology}
}
\affiliation[ln]{organization={Lingnan University}
}
\affiliation[CAIR]{organization={Centre for Artificial Intelligence and Robotics (CAIR) Hong Kong Institute of Science \& Innovation, Chinese Academy of Sciences}
}
\affiliation[thu]{Tsinghua University}

\affiliation[JX]{Fujian Jiangxia University}

\begin{abstract}
The explosive growth of a series of AI-driven applications, software and tools (AI entities) led by the flourish of Large Language Models (LLMs) and other generative algorithms have caused an explosive increase in the number of AI applications on the internet. The data from interactions between AI applications and users is growing increasingly, and the resulting security concerns and ambiguities in responsibility attribution have become critical challenges that need to be addressed urgently. Efficient means of traffic recording are required to support dynamic review and auditing of the behavior and usage of AI applications to prevent and trace back to the source of accidents. In this paper, We propose a blockchain-based, AI usage AAT traffic recording and governance framework called AiAuditTrack (AAT), which aims at filling the AAT gaps required for risk traceability and auditing. AAT builds an identity management mechanism centered around decentralized identity (DID) and verifiable credentials (VC) to achieve the construction of identifiable and trusted authentication identities for AI entities, and records the interaction trajectories between entities on the chain through identity identifiers, achieving cross-system and cross-module AI entity supervision. The AI entities are viewed as nodes in a complex graph, with edges represent the interaction trajectory at a specific point in time. Furthermore, a risk diffusion algorithm is designed to effectively trace the source of risky actions and issue diffusion warnings to other AI entities involved. 
To demonstrate the operational capacity of AAT, Transactions Per Second (TPS) data of blockchain are used to examine the credibility and stability of AAT in terms of supporting a large volume of interaction AAT recording. Our work on constructing AI interactive networks can provide insights for both research and industry, analyzing their topology and key nodes using complex system theory and risk propagation dynamics; our AI entity identity and behavior recording mechanism supports an interactive audit system with traceability, verifiability, and scalability. By introducing trajectory encoding and risk diffusion models, we provide structured risk management and responsibility tracing solutions for a complex concelebration environment, providing a practical path and technical foundation for building a trustworthy and secure environment.

\end{abstract}
\end{frontmatter}

\section{Introduction}
Large Language Models (LLMs) have achieved revolutionary breakthroughs, demonstrating unprecedented language understanding, reasoning, and knowledge integration capabilities. Through conversations or commands in natural language, users can directly interact with applications without investing time in professional training. Driven by the innovations, the number of AI applications has experienced explosive growth; meanwhile, when AI applications evolve from closed text generators into agents capable of proactively executing tasks—even when multiple agents collaborate to perform work—this gives rise to increasingly complex communications between AI applications and a massive volume of generated content across the internet\cite{zhang2023large,lu2022survey,shen2024llm,fan2024survey,xi2025rise,hu2021decentralized,li2024survey}.

Generated content inevitably exhibits hallucinations (untrue information) and systemic biases due to the probabilistic nature of AI algorithms and the inherent biases in training data. Despite the fact that academia and industry have employed numerous sophisticated methods to minimize the probability of generating such harmful information, malicious actors can still deliberately create harmful content through means like “jailbreaking”.\cite{rae2021scaling,kumar2025no,shi2024large,zhuo2023red,kamath2024llm,huang2025survey,zhang2025siren}. 

Moreover, the complex interaction patterns among multiple AI applications further expand the system's attack surface, making it difficult for security specification technologies targeting single models to ensure the safety of the entire system's collaborative outcomes. Examples include tool poisoning attacks where malicious instructions are embedded in tool descriptions, and unauthorized malicious calls in agent interactions\cite{ehtesham2025survey,yu2025survey,wang2025sok,zhang2024agent}.

Adding to aforementioned issues, tracing the responsible parties after incidents occur has become even more challenging. This is because there is no unified platform for storing and managing the information regarding calls and interactions between AI applications; currently, interaction records are stored independently by the operators of each application. As a result, it is difficult to integrate credible cross-domain data for auditing when incidents arise.

Previous solutions such as Model Context Protocol (MCP) and Agent-to-Agent Protocol (A2A) standardize the interaction protocols for AI applications\cite{ehtesham2025survey,yu2025survey}, enabling different applications to comply with the unified set of security guidelines. However, they are unable to address targeted systemic attacks. The standardized processes led by industry norms and community initiatives have reached the limit of security prevention, necessitating more tangible cross-platform management, control, and analysis of interaction data between AI applications to effectively implement more precise, dynamic and efficient security strategies.

At present, while LLM and LLM based systems are widely applied, the hidden security threats of LLM systems cannot be completely solved. How to regulate and trace the artificial intelligence entities that cause harm to LLM and systems to prevent large-scale harm is an important issue now.
We propose an AI governance framework called AiAuditTrack (AAT), which aims to record the traffic of AI-driven applications, support pre- and post-incident audit of interaction trajectories, and dynamically infer risk warnings and responsibility tracing in interactions. In this study, our main contributions are: (1) proposing an AI governance framework for root cause analysis of accidents in AI interactions; (2) Our study describes the interaction trajectory between AI applications from perspective of blockchain and graph structure, introduce a dimension for handling AI interaction problem; (3) proposing a risk warning method in AI interactive networks to prevent the spread of accidents that have already occurred.

\section{Related Work}
\subsection{AI Security}
Both academia and industry have conducted extensive researches and developed various approaches on AI security governance.

Google proposes a secure AI framework that aims to ensure the security and reliability of AI systems through multilevel security measures and governance mechanisms\cite{GoogleSAIFApproach}.
\cite{xia2024ai}A framework for evaluating AI systems has been proposed, emphasizing the importance of shifting from model evaluation to system-level evaluation.
\cite{liang2024design}Propose a AI security audit management system framework based on user entity behavior analysis (UEBA) and integration of AI technology, to enhance the detection capability of complex and internal threats.

The National Telecommunications and Information Administration (NTIA) of the United States proposed a systematic AI accountability framework in its 2024 AI Accountability Policy Report, highlighting the use of an "accountability chain" model that integrates information flow, independent evaluation, and accountability mechanisms to promote the construction of reliable AI \cite{AI_Accountability}.

\subsection{Blockchain, Identity and Verification}
With the development of identity identification standards, trustless identity authentication mechanisms have become a hot topic. In this context, A Novel Zero Trust Identity Framework for Agenetic AI proposes a complete identity management framework for AI agents. This system combines DID, verifiable credentials (VC), agent naming services (ANS), and fine-grained access control based on zero knowledge proof (ZKP), supporting cross domain and cross chain agent registration and authentication\cite{huang2025novel}.

LOKA Protocol propose the "Universal Agent Identity Layer (UAIL)", which not only includes DID/VC technology, but also integrates a decentralized ethical consensus mechanism, enabling agents to have a trusted identity and moral controllability\cite{ranjan2025loka}.
Non-Disclosing Credential On-chaining for privacy-preserving authentication leverages zk-SNARK to support authentication process under the premise of no information leakage, covering application scenarios with limited sensitive data\cite{heiss2022non}.
Verifiable AI proposes an identity and behavior verifiable framework for AI agents and AI generated content, ensuring the authenticity of content sources through on chain records\cite{Verifiable_AI}.
\cite{article} proposes a decentralized, immutable, and transparent method to protect user identity. By integrating blockchain with AI, organizations can achieve secure identity verification, AAT integrity, and trustless interaction between users and intelligent systems.

The Agentic AI security framework proposed by Oracle also points out that in the post zero trust era, the identity and behavior logs of intelligent agents should be bound through blockchain mechanisms to ensure the auditability of behavior\cite{ladd2025beyond}.
The decentralized AI agent communication framework integrates decentralized authentication using decentralized identifiers (DID), hybrid encryption, dynamic threat detection, and audit logs to build a secure communication framework between AI \cite{Decentralized}.

The Agent Communication Protocol is based on the unique identification system of Agent Identity, which ensures the uniqueness and traceability of each agent in the network. It also uses end-to-end encryption and digital signature technology to ensure the security and integrity of communication between agents.\cite{ACP_WhitePaper}

\section{AIAuditTrack Framework}
In this study, we introduce a novel AI security framework based on blockchain, AIAuditTrack (AAT), for positioning individuals who generate incorrect or harmful information in the complex and intricate interaction of artificial intelligence applications, then deliver the risk to nodes that interact with accident related nodes, to address the problem of responsibility attribution for accident and risk prevention. 

\subsection{Background and Notations}
We consider the interaction between AI entities as a complex graph and design a series of processes for failure responsibility positioning and risk diffusing. Given AI entities ($V$), each entity ($v$) is denoted as a specific node from the perspective of graph. Each $v$ has an on-chain address which serves as unique decentralized identifier (DID). Through DID, the description file (DID Document) of each node can be indexed to obtain the Verified Credentials (VC), entity feature descriptions (e.g. model card), entity risk level and other designated information stored on or off chain, depending on specific system design requirements and implementations.

When one entity ($v_i$) interacts with another ($v_{j}$), a directed data flow (edge) is formed. Thus, the interaction between entities is denoted as $\delta(v_{i}, v_{j})$, which represents the interaction traffic from one node to another. 

In scenario where collaboration of agents, AI applications and tools are involved, multiple interactions belonging to the same task could therefore form a trajectory, denoted as $E_{x} = \{\delta_{0}, \delta_{1}, ..., \delta_{k}\}$, which characterizes the sequence of interactions between entities over time. To facilitate the querying and auditing of the behaviors and relationships of AI entities in the later stage of AAT system, each trajectory is assigned with a globally unique trajectory ID denoted as $x$. The topological structure of these occurred interactions and invocation relationships is described as $T = (V, E)$, where $E$ represents the set of all edges (interactions).

A table describing the notations used in this paper is provided in \ref{tab:notations}.

\begin{table}[H]
    \centering  
    \begin{tabular}{p{4cm}p{9cm}}  
      \toprule
      \textbf{Symbol} & \textbf{Meaning} \\
      \midrule
      $V$  & AI entities represented as a set of nodes on chain \\
      $v_i$  & A specific AI entity from $V$ \\
      $\delta(v_{i}, v_{j})$ & Edge from node $v_{i}$ to node $v_{j}$ \\
      $E_x$ & A trajectory of interactions between AI entities (a sequence of edges) \\
      $E$ & Set of all edges (interactions) in the graph \\
      $x$ & Trajectory ID \\ 
      $T$ & Topological structure of trajectory and entities. \\
      \bottomrule
    \end{tabular}
    \caption{Notations}  
    \label{tab:notations}  
\end{table}

\subsection{Trajectory Record}
\label{TrajectoryRecord}
When a user initiates a call request to an AI entity denoted as $v_0$,—as the initial receiver in this trajectory—$v_0$ will first request a Trajectory ID ($x$) after verifying the identity of $v_0$. 

Upon obtaining the trajectory ID, $v_0$ transmits records of receiving the user's request, forming the first interaction record of this trajectory. From AI entity's perspective, the trajectory ID, sender's DID (in this case, the sender is none) and receivers' DIDs will appear as a "transaction record" under AI entity's on-chain address. 

If the task requires the collaboration of multiple AI Entities ($v_1, v_2, ..., v_n$), the initial AI entity $v_0$ will pass this trajectory ID down to the subsequent entities, and these entities will update their own records with the same trajectory ID to the blockchain under their on-chain address.

Accordingly, for the interaction trajectory generated by the same task, each AI entity's address shall maintain corresponding inbound and outbound data traffic records under their on-chain address. The records across all AI entities involved in the same task are cross-verified against one another, thereby fulfilling the objective of mutual reconciliation.

\begin{figure}[!htb]
    \centering
    \includegraphics[width=0.9\textwidth]{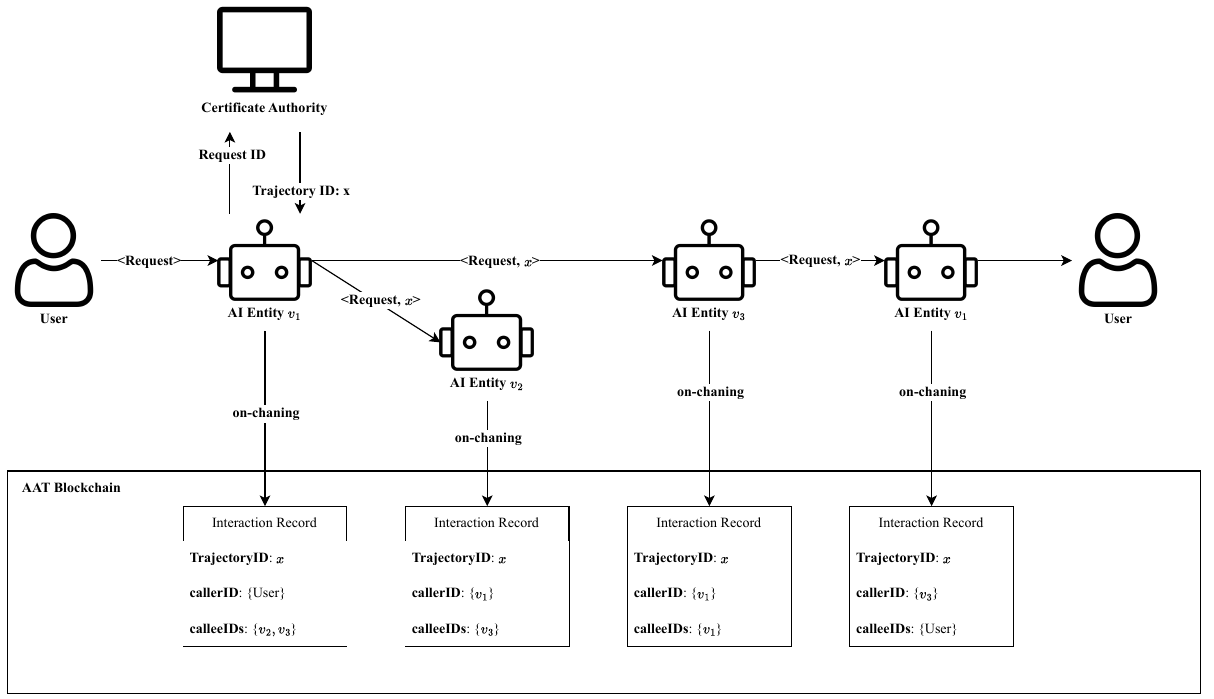}
    \caption{Trajectory Record}
    \label{E_create}
\end{figure}

At a minimum, a traffic record stored on the blockchain should contain the following content:$<callerID, calleeIDs, TrajectoryID>$. Here,$callerID$ refers to the ID or on-chain address of the initiator of this specific interaction, with the specific type (ID or on-chain address) depending on the practical design and requirements of the system; $calleeIDs$ denotes the set of receivers involved in this interaction; and $TrajectoryID$ represents the ID of the task. The process as \ref{E_create} show. 


\subsection{Trajectory History Restoration}
When an incident occurs, the system can restore the relevant topology graph by combining external data from the blockchain network (such as timestamps, watermarks, etc.). Thus, a subset graph encompasses all AI entities related to the incident as well as the data traffic trajectories between these AI entities. This graph structure can furnish data support for subsequent traceability, auditing, and other analytical endeavors.

Each AI entity accident or anomaly corresponds to a set of unique trajectory $E_x$, and AAT marks the set of risk nodes $V_{inspect}$ based on the trajectory code $x$ and the DID of the AI entity.
For each $v_r$ in $V_{suspect}$, the interaction records in AATChain corresponding to its DID together constitute $T_{merge}$.

AAT will traverse $V_{suspect}$ in AATChain according to the trajectory $E_x$ in $T$. During this process, evidence will be reviewed for each $v_r$, and finally marks the responsible party $V_{risk} $for the accident. Specifically, when AAT receives a risk accident alert, it will locate the code of all running trajectories with the accident time interval as the core index, and reproduce the complete trajectory within the interval through the code and records on AATChain. 

Subsequently, based on the specific type, nature, and scene characteristics of the accident (such as watermarks), AAT will mark $V_{inspect}$ and associated trajectories to narrow down the audit scope and focus on the critical traceability path for nodes on the critical traceability path, AAT will perform authentication operations on the DID documents of these nodes within the accident time interval. Extract key information such as node qualifications and real-time operational status through DID documents, and identify potential accident liability parties.

\subsection{Risk Warning Diffuse}
We have defined a risk level update process ERLdiffuse for the risk audit process to update the node $V_{infection}$ that directly or indirectly interacts with $V_{risk}$ of $T_{merge}$, as shown in \ref{ERLdiffuse_process}. 

ERLdiffuse is a diffusion process, specifically, with $V_{risk}$ as the central initial node. Each node $N(v_{r})$ that $v_r$ interacts with directly calculates the updated $ERL_{update(N(v_{r}))}$ through $ERLdiffuse(ERL_{v_r},ERL_{N(v_{r})},{\beta}_{v_i})$. Based on these $N(v_{r})$ DIDs, the ERL in its archive is updated through the information change process. The calculation of $ERL_{update(v)}$ is determined only by the properties of the adjacent nodes and the node itself. If only considering the ERLdiffuse of risk nodes and security nodes, then in the process of risk level transmission, the calculated $ERL_{update(N(v_{i}))}$ is usually greater than $ERL_{update(N(N(v_{i})))}$.

If $ERL_{update(v)}$ is 0, the ERLdiffuse process at this node is terminated. Therefore, when there is no node to transfer the risk level, the ERLdiffuse of the entire process ends.

\begin{algorithm}
\caption{ERLdiffuse Process for Risk Level Update}
\label{ERLdiffuse_process}
\begin{algorithmic}[1]
\State \textbf{Input:} $V_{risk}$, $T_{merge}$, $V_{infect}$, $\beta_{v_i}$
\State \textbf{Output:} Updated ERL values for infected nodes

\State \textbf{Initialize:} Set $V_{risk}$ as initial active nodes
\State \textbf{Process:} Update risk levels for nodes interacting with $T_{merge}$

\While{Active nodes exist for risk propagation}
    \For{each risk node $v_r \in V_{risk}$}
        \State Get directly connected neighbors $N(v_r)$
        \For{each neighbor $v_i \in N(v_r)$}
            \State Calculate updated ERL: 
            \State $ERL_{update(v_i)} \leftarrow ERLdiffuse(ERL_{v_r}, ERL_{v_i}, \beta_{v_i})$
            \State Update node $v_i$'s ERL in its profile based on DID
            \State $ERL_{update(v_i)}$ depends only on adjacent nodes and node's own attributes
        \EndFor
    \EndFor
    
    \State \textbf{Propagation Rule:} 
    \State Considering only risk and safe nodes:
    \State $ERL_{update(N(v_i))} > ERL_{update(N(N(v_i)))}$ in risk level transmission
    
    \State \textbf{Termination Condition:}
    \If{$ERL_{update(v)} = 0$}
        \State Stop ERLdiffuse process at node $v$
    \EndIf
    
    \If{No nodes propagate risk levels}
        \State Terminate entire ERLdiffuse process
    \EndIf
\EndWhile

\State \textbf{return} All updated ERL values
\end{algorithmic}
\end{algorithm}

\subsection{Profile Updates}
The description files of AI entities often require information changes. In AAT, the scenario where information changes occur in archives is described as follows: the risk level of the AI entity is adjusted after a risk audit, or the risk level of the AI entity needs to be adjusted after rectification and liquidation; The issuing unit of an AI entity needs to update its $\beta$ due to changes in its feature description after iterative updates, or changes in its domain standards.

\section{System Design}
\begin{figure}[!htb]
    \centering
    \includegraphics[width=0.9\textwidth]{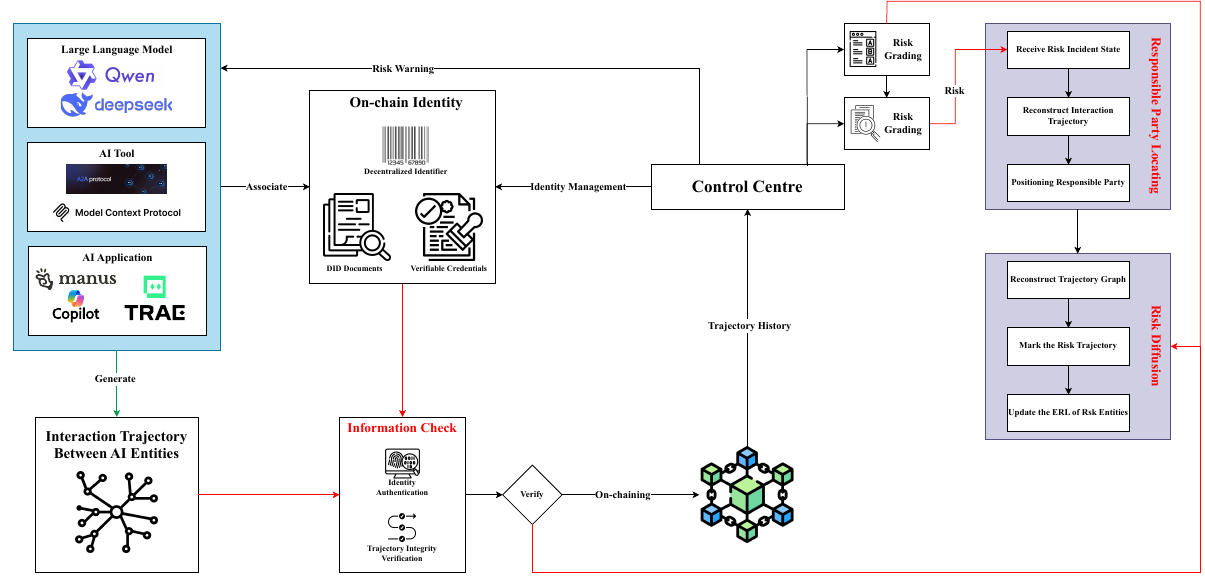}
    \caption{System Workflow}
    \label{system}
\end{figure}

In this part, we construct an AAT system based on AAT framework and implemented an engineering solution for it, as shown in \ref{system}. We will discuss in detail the construction and collaborative work of various parts of the AAT system to achieve coupling with the AAT framework workflow. We have provided a specific system implementation solution, which can be viewed in the \ref{engineeringsolution}. The foundation of the artificial intelligence interactive supervision system is a blockchain system , a control center within supervision system receiving information from external environment or internal module and coordinating the work of various functional modules. The functional modules are divided into identity management module, risk diffusion module and fault traceability-positioning module. We have introduced Risk Grading and Risk Discover to construct scenarios for detecting fault risks. 

The major design objectives of the system are as follows: verifiable identities (DID and VC) are assigned to each AI node; trajectories are recorded; interaction graphs are constructed; and rapid traceability, as well as risk diffusion and control, are achieved when risky behaviors occur.

\subsection{Overview of Architecture}
In general, the AAT system incorporates an on-chain and off-chain collaborative components, which are divided into three layers overall:

\paragraph{\textbf{On-chain Layer}} 
Smart contracts written in Move implement functions such as identity management, VC registration and approval, on-chain recording of call behaviors, and risk level recording, with identity verification conducted before recording call events.

\paragraph{\textbf{Off-chain Analysis Layer}}
It queries on-chain event data in real-time and dynamically generates call trajectory graphs when requests are initiated, which are used for traceability and risk analysis.

\paragraph{\textbf{Audit Layer}}
It provides interfaces for functions such as identity information query, call path visualization, and risk level checking.

\subsection{Identity Control and VC Management Module}
 In AAT, each artificial intelligence entity will be endowed a verifiable identity.  AAT system leverages identity management module to construct and modify the identity of entities. When a new artificial intelligence entity enrolls in the system, identity management module will assign verifiable identities (DID + VC) and construct DID document to it. The DID document content is shown in \ref{DIDdocument}. As issuing a modification request in the control center, identity management module will modify the DID document of corresponding entities, such as entity itself undergoes functional iteration or ERL has changes after an accident occurs.

In AAT, the address of each AI entity on the Sui blockchain is employed as its unique identity identifier (i.e., DID). Metadata can be registered by each node via on-chain contracts, and applications for VCs can be initiated, which are subsequently issued by the preset approving authority (i.e. CA).

The table \ref{tab:contractInterfacesIDandVCDocument} shows the design of interface for identity and VC document management. The VC documents is generated and submitted by the producer or manager of AI entities in the form of JSON texts. These documents contain information encompassing identity attributes, issuance criteria, validity periods, and other critical details for CAs to evaluate.

When initiating the credential process, the AI entity's manager first registers basic information through the \texttt{register\_metadata(account: \&signer, metadata\_uri: String)} interface, which records the metadata URI associated with the entity's identity on the blockchain. Following this, the applicant submits the VC application via the \texttt{apply\_vc(account: \&signer, vc\_hash: vector<u8>, vc\_uri: String)} interface, where the \texttt{vc\_hash} parameter represents the cryptographic hash of the JSON document, and \texttt{vc\_uri} points to the storage location of the original document (such as a distributed file system or secure server).

Upon review, the CA approves valid applications using the \texttt{approve\_vc(issuer: \&signer, applicant: address)} interface, which triggers the on-chain storage of the VC hash and URI. Conversely, invalid applications are rejected through the \texttt{reject\_vc(issuer: \&signer, applicant: address)} interface, with the rejection recorded immutably on the blockchain. A similar process using these interfaces is repeated for each subsequent update of the VC document. This approach ensures that sensitive content is prevented from being exposed during on-chain transmission, while subsequent verification of data integrity can be achieved through hash comparison.
\begin{table}[!htb]
    \centering
    \caption{Proposed Interfaces for Identity Control and VC Management}
    
    \begin{tabular}{p{0.6\textwidth}p{0.3\textwidth}}
        \toprule
        \textbf{Interface Definition} & \textbf{Description} \\
        \midrule
        \texttt{public entry fun register\_metadata(account: \&signer, metadata\_uri: String);} & Registers metadata for a node through the signer's account, with metadata address specified via URI. \\
        \midrule
        \texttt{public entry fun apply\_vc(account: \&signer, vc\_hash: vector<u8>, vc\_uri: String);} & Initiates a VC application by the node, including hash and URI of the credential content. \\
        \midrule
        \texttt{public entry fun approve\_vc(issuer: \&signer, applicant: address);} & Approves the VC application by the designated issuer (CA), validating the applicant's credentials. \\
        \midrule
        \texttt{public entry fun reject\_vc(issuer: \&signer, applicant: address);} & Rejects the VC application by the designated issuer (CA), marking the application as invalid. \\
        \bottomrule
    \end{tabular}
    \label{tab:contractInterfacesIDandVCDocument}
\end{table}

\subsection{Trajectory Management Module}

When recording interactions, if the interaction is the first in a trajectory, the CA will assign a globally unique trajectory ID $x$. The AI entity who initiated the interaction will record this ID together with its DID and the designated receivers' address to the blockchain.

\begin{table}[!htb]
    \centering
    \label{tab:}
    \caption{Proposed Interfaces for Trajectory Management Module}
    \begin{tabular}{p{0.6\textwidth}p{0.3\textwidth}}
    \toprule
    \textbf{Interface Definition} & \textbf{Description} \\
    \midrule
    \texttt{public entry fun log\_interaction(sender: \&signer, callee: address, trajectory\_id: vector<u8>, action\_type: string, interaction\_hash: vector<u8>);} & Save the call records to the chain \\
    \bottomrule
    \end{tabular}
\end{table}

The InteractionEvent is a struct with drop and store attributes, designed to record detailed information of behavioral interaction events on the chain. It comprises six fields: caller (the address of the caller), callee (the address of the callee), trajectory\_id (the trajectory identifier stored in the form of a byte array), action\_type (a string indicating the type of behavior), interaction\_hash (the interaction hash value stored as a byte array), and timestamp (a 64-bit unsigned integer recording the time when the event occurs). These fields collectively form a complete record of an on-chain interaction event, clearly presenting the participating entities, the associated trajectory, the nature of the behavior, the unique identifier, and the time information.
\subsection{Trajectory Construction Module}
When users or auditing parties need to audit the records of interactions or data traffic between AI entities, this module traces back the relevant AI entities and interaction edges based on the specified trajectory ID, thereby visualizing directed graph of the trajectory. This graph serves to support interaction visualization, traceability analysis, and risk propagation control:

\begin{enumerate}
    \item Initiate a query request: e.g., \texttt{GET /trajectory/:trajectory\_id}
    \item On-chain data query: Use the Sui GraphQL API to retrieve all \texttt{InteractionEvent} matching the \texttt{trajectory\_id}.
    \item Graph structure generation: Each event is constructed as a \texttt{caller $\to$ callee} edge to form the graph \( G = (V, E) \), where \( V \) represents vertices (AI entities) and \( E \) represents edges (interactions).
    \item Graph analysis operations: Support source node tracing (reverse BFS) and propagation path analysis (forward BFS).
\end{enumerate}

\begin{figure}[!htb]
    \centering
    \includegraphics[width=0.9\textwidth]{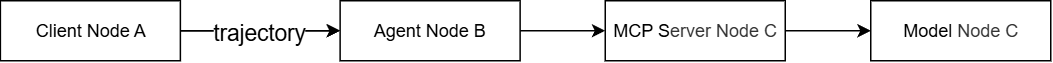}
    \caption{Trajectory Record Construction}
    \label{trajectoryRecordConstruction}
\end{figure}

\subsection{Risk Level Update Module}
When abnormal behavior of an entity is detected, auditors can submit a risk report. The system will analyze the propagation path based on the trajectory graph of the node and automatically update the risk levels of relevant nodes.

The system is capable of constructing a trajectory graph, which delineates the entities involved in the interaction process. The risk levels of the associated entities shall be adjusted in accordance with the Risk Warning Diffusion mechanism as defined in Section 3.4.

\section{System Analysis}

In this chapter, there is an analysis and validation will be conducted in the system of Artificial Intelligence Security Supervision System based on AAT Framework encompassing two dimensions: the verification of on-chain identity trustworthiness, the traceability verification of AI interaction behaviors. And the load analysis of blockchain for artificial intelligence interaction nodes will show in \ref{load analysis}.

\subsection{Experimental environment}
A DID and VC verification environment was set up, with the W3C standard DID document structure and VC model\cite{VCmodel} used for the on-chain identity trustworthiness verification experiment of AAT.

The experiment included a centralized identity management center, representing the previously described CA for issuing credentials, and four types of AI nodes: legitimate AI entities simulating the holding of valid VCs, and illegal nodes attempting to launch attacks. The verification process relied on the DID documents and public keys stored on the chain to perform signature verification, revocation list queries, and VP/VC binding verification. All attack attempts were simulated through interface calls, with real-time verification responses recorded.

For the traceability verification of AI interaction behavior, a simulated multi-node interaction platform was built, which included four types of nodes: LLM nodes, decision module nodes, AAT processor nodes, and multi-computing platform (MCP) nodes. Sequential interactions (A $\to$ B $\to$ C $\to$ D) and parallel interactions (A $\to$ \{B1, B2\} $\to$ C) were initiated through scripts, and all interactions were recorded on the chain via smart contracts with metadata (timestamps, call relationships, parameter hashes, etc.). In the experiment, some nodes were set to states of "interruption", "tampering", or "information missing" to test the accuracy of the system's full path backtracking and branch recognition.

\subsection{Blockchain Identity Trustworthiness Verification}
We designed an experimental analysis to ensure that the DID and VC held by AI nodes in AAT can be securely registered, issued, verified, and have anti tampering capabilities: (1) Build a verification module that supports VC format and DID document parsing\cite{DIDs,DIDsMethod}, and simulate the verification end; (2) The centralized identity management center uses a private key to issue and register DID to the chain for VC; (3) Construct four typical types of attack VC data that include illegal tampering, misuse, replay, forgery, and other behaviors; (4) Perform VC verification process, record whether each verification passes, time taken, and attack identification results. We use table \ref{reliability} indicators to quantify the experimental results\cite{salman2018security,mazzocca2025survey}.
\begin{table} 
    \centering  
    
    \begin{tabular}{p{4cm}p{4cm}p{4cm}}  
      \toprule

      Indicator & Definition & Standards \\ 
 
      \midrule
      Integrity Validation & The proportion of legitimate VC signatures that pass verification & ISO/IEC 27001 A.9.2.1 \\ 
      Tamper Rejection & The proportion of cases where illegal modifications to VC or DID are identified and blocked & NIST SP 800-53 SI-7 \\  
 
      \bottomrule
    \end{tabular}
    \caption{identity trustworthiness verification indicators and standards}  
    \label{reliability}  
  \end{table}

\subsection{Traceability Verification of AI Interaction Behavior}
We evaluate the integrity, recoverability, and anomaly detection capability of AAT interaction records. Specifically, we preset functions and identity credentials for each type of node, arrange simulated multi hop and multi branch task flows, write all tasks onto the chain through contracts, establish an interaction graph, set specific nodes as "abnormal nodes", tamper with interaction information or skip task reporting, use a traceability engine to backtrack the task path from the final output, and evaluate the accuracy of path recovery and node recognition. We use table \ref{tab:Traceability verification of AI interaction behavior} indicators to quantify the experimental results\cite{ahmad2018towards}.

\begin{table}[htbp]  
    \centering  
    \begin{tabular}{p{4cm}p{4cm}p{4cm}}    
      \toprule
 
      Indicator & Definition & Standards \\  
 
      \midrule
      Audit Trail & The proportion of on chain records of interactions between nodes & ISO 27001 A.12.4 \\  
      Attribution Accuracy & successfully trace back to the source node and the responsible node & NIST AU-6 (Audit Review) \\  
      Multi-hop Traceability & the complete path be restored in a multi-step interaction scenario & ISO/IEC 27035 (Incident Response) \\  
      Path Fork Accuracy & Accuracy of path recognition for concurrent/merged nodes & Provenance Systems Practice \\  
      \bottomrule
 
    \end{tabular}
    \caption{Traceability verification of AI interaction behavior}  
    \label{tab:Traceability verification of AI interaction behavior}  
  \end{table}

\subsection{Results}
The results presented in Table \ref{result} were obtained from our experiments on on-chain identity trustworthiness verification and AI interaction behavior traceability. Our experimental analysis indicates that AAT exhibits high integrity and robustness in its identity verification mechanism, and is capable of fully recording and restoring multi-hop and concurrent interaction paths of AI nodes. It also demonstrates good trajectory integrity and path recognition capabilities in composite task scenarios.

Typical attack simulation scenarios were conducted on AAT, with effective interception achieved in all four typical attack scenarios, as shown in Table \ref{defense}, which meets international identity security standards. The system verification efficiency meets practical application requirements, indicating that AAT can effectively identify and prevent identity forgery, tampering, and replay attacks, thereby ensuring the credibility of on-chain identity information.

\begin{table}[htbp]
    \centering
    \begin{tabular}{p{4cm}p{2cm}p{4cm}p{2cm}}
    
      \toprule
       
      \multicolumn{2}{c}{\textbf{trustworthiness verification}} & \multicolumn{2}{c}{\textbf{Traceability verification}}  \\  
       
      \textbf{indicator} & \textbf{result} & \textbf{indicator} & \textbf{result}  \\
      \midrule
      Integrity Validation & 100\% & Audit Trail& 100\%\\
      Tamper Rejection & 100\% & Attribution Accuracy & 100\%  \\
      -- & -- & Multi-hop Traceability & 100\% \\
      -- & -- & Path Fork Accuracy & 100\% \\
    
      \bottomrule
       
    \end{tabular}
    \caption{result}
    \label{result}
  \end{table}

\begin{table}[htbp]  
    \centering  
    \begin{tabular}{p{4cm}p{4cm}p{3cm}}  
      \toprule
        
      type & method & defense  \\  
        
      \midrule
      VCforgery & Tamper with VC and falsifying signatures & DID Validator+Issuer key verification \\  
      VCtransfer & Impersonator use VC,VP & VP binding DID \\  
      Replay attack & submit intercepted VC & nonce+time stamp and status verification \\  
      Revocation use& Attempt to resubmit after being revoked VC & Revocation Registry \\  
      \bottomrule
        
    \end{tabular}
    \caption{defense}  
    \label{defense}  
  \end{table}

\section{Conclusion}
In this article, we propose the AiAuditTrack (AAT) framework for AI application interaction traceability auditing, aimed at addressing the security challenges and accountability issues faced by current AI systems with large language models at their core in complex interactions. This framework combines the auditability and immutability features of blockchain, and designs an intelligent agent identity construction mechanism, an interaction trajectory on chain mechanism, and a risk tracing and diffusion mechanism. It can effectively identify the responsible party and control the spread of accident risks in the event of anomalies or harmful content generation in AI systems.

Our design focus includes: (1) building AI entity identities through DID and VC, achieving on chain identity management and identity authenticity verification; (2) By utilizing trajectory unique encoding and node recording mechanisms, the traceability and tracking of complex interactive behaviors between AI applications can be achieved; (3) Propose a risk diffusion algorithm based on risk level propagation to provide a computable risk control mechanism for AI systems. (4) Viewing AI interactive networks as complex networks, one can draw on complex systems theory to analyze their topology, key nodes, and risk propagation dynamics. Our research can provide infrastructure support for the governance, security, and transparency of AI systems.

The characteristics of AAT were analyzed, and directions for future research were organized: (1) Performance and Scalability Optimization: As a proprietary blockchain system, AATChain may have performance bottlenecks when facing high-frequency AI interactions. In the future, lightweight consensus mechanisms can be introduced or Layer2 schemes can be adopted to improve write efficiency and scalability.
(2.) On chain and off chain collaborative data: Develop cross chain data merging algorithms to interact with external chains or off chain data (such as intelligent agent alliance chains, public certificate chains), ultimately improving the business adaptability, data credibility, collaborative efficiency, and ecological scalability of proprietary chains.
(3.) Introduction of intelligent agent behavior anomaly detection model: Currently, AAT relies on interaction trajectories and rules for risk tracing. In the future, methods such as graph neural networks can be introduced to construct anomaly detection models for interaction graphs, achieving active identification of potential risk nodes.
(4.) Integration and evaluation with actual platforms: 

In the future, we plan to deploy AAT prototype systems in open-source Agent systems or enterprise AI platforms for large-scale experiments to evaluate their accuracy, real-time performance, and governance effectiveness.
(5.) Expansion of Compliance and Ethical Mechanisms: Combining existing AI ethical guidelines and compliance policies, further enriching the risk level update mechanism, introducing ethical factors and contextual constraints, and enhancing the identification and governance capabilities of malicious agents.

\bibliographystyle{elsarticle-num} 
\bibliography{main}

\newpage  
 
\appendix
 
\section{Blockchain load analysis}\label{load analysis}
In order to analyze the operational load of AAT in high-frequency request scenarios of AI systems, we collected key data from public chains and AI platforms for analysis and demonstration: we collected Qubic's maximum TPS performance indicator data and the daily active user numbers of DeepSeek and OpenAI, and set the per capita daily request volume to 100 times; We calculate the total daily interaction volume and convert it into the required TPS on average, and finally calculate the redundancy ratio of Qubic TPS to the required TPS.

Qubic, a representative high-performance public chain in the industry, was adopted as the performance baseline. Combined with daily active users  from OpenAI and DeepSeek, the daily interaction volume was estimated by calculating user request behaviors. Subsequently, the required TPS (transactions per second) of the system was calculated, and comparisons were made with Qubic's theoretical TPS value to evaluate its support capability in massive AI interactions.

Data were collected showing that the maximum TPS of Qubic is 15.5 million\cite{Qubicperformance,Qubicreport}, while the daily active users of DeepSeek and OpenAI are 2.2 million and 1.2 million\cite{DeepSeekStatistics,ChatGPTStatistics}, respectively.

Thus, under the premise of 100 AI requests per person per day, with a total of 2.2 billion to 12 billion requests per day, the estimated average TPS required for AAT is 25,463 to 138,889.

If estimated based on Qubic, the simulated TPS redundancy ratio of AATChain is 110, indicating that blockchain still has good performance carrying capacity in high-frequency AI interaction scenarios.The data details are shown in table \ref{Load estimation}
\begin{table}[htbp]  
    \centering  
    \begin{tabular}{p{4cm}p{4cm}p{3cm}p{3cm}}  
      \toprule
       
      term & definition & data & source \\  
       
      \midrule
      QubicmaxTPS & peak value & 15,500,000 TPS & Qubic Measured data\\  
      DeepSeekactivity & Daily average number of active users & 22,000,000  & DeepSeek data\\  
      OpenAIactivity & Daily average number of active users & 120,000,000  & OpenAI data\\  
      Total daily requests & Estimation of 100 AI requests per person per day & $\geq$2.2B $\sim$ $\geq$12B& Estimation calculation: Number of users x 100 requests per person\\  
      average TPS & Throughput required for continuous service requests & 25,000 $\sim$ 140,000 TPS & Estimation calculation: Total daily requests/86400\\ 
      TPSredundancy & Qubic TPS / requiredTPS & 110$\times$ $\sim$ 600$\times$ & Redundancy estimation\\  
      \bottomrule
       
    \end{tabular}
    \caption{Load estimation}  
    \label{Load estimation}  
  \end{table}

\section{System Engineering solution}\label{engineeringsolution}
This section provides a detailed introduction to an AI Audit Trail system implementation plan, including the overall architecture, module functions, contract interface design, and off chain processing flow. The system is implemented on the Sui blockchain platform, using Move language to develop core contracts, and combining off chain services for call trajectory construction and risk analysis. The system design objectives include: assigning verifiable identities (DID+VC) to each AI node, recording call behavior trajectories (Trajectory), building a call network graph, and achieving rapid traceability and risk propagation control in the event of risky behavior.

\section{DID document}
\begin{table}[H]
    \centering  
    \begin{tabular}{p{4cm}p{9cm}}  
      \toprule
      \textbf{item} & \textbf{Meaning} \\
      \midrule
      Basic identifier & The unique decentralized identifier (DID) of each AI node on the chain, which is its "ID number" in the entire AAT system. \\
      Node Type  & Identify the specific type of AI applications, such as AI Agent, Large Language Model, Model Context Protocol Server, Data Processing Module, Decision Support System, Sensor Interface, etc. This helps to understand its role in auditing stage. \\
      Functionality & A detailed description of the functionality of a AI application, such as "image generation", "text summarization", "data analysis", "risk assessment", "user interaction interface", etc.  \\
      Owner/Operator & The on chain identity DID associated with the enterprise or institution that owns or is responsible for the AI application. This establishes a verifiable linkage between AI nodes and actual responsible parties. \\
      Version information & Record the version of the AI application. This is crucial for tracing the risks caused by specific versions, even if the model is subsequently updated or replaced, historical version behaviors can still be traced. \\ 
      Capability Statement & Declare the specific capabilities, qualifications, or certifications that an AI application possesses in the form of verifiable credentials (VC). For example, a VC can prove that an AI application meets certain industry standards, possesses the qualifications to provide certain services, or has passed particular audits. Depending on circumstances, a VC may have a valid period or condition. VCs are issued by authoritative institutions such as management centers or certification bodies. \\
      ERL & The initial risk assessment and risk level is issued by CA when AI application initially registered on AAT system. Subsequent changes in the risk level are dynamically adjusted based on its interactive behaviors during operation and audit results. \\

      \bottomrule
    \end{tabular}
    \caption{DID document}  
    \label{DIDdocument}  
\end{table}

\end{document}